\documentclass[runningheads,a4paper]{llncs}
\usepackage[english]{babel}
\usepackage[utf8x]{inputenc}
\usepackage{graphicx}
\usepackage{url}
\usepackage{hyperref}
\usepackage{caption, comment}

\title{Algorithmic Songwriting with ALYSIA}

\author{}
\institute{}

\author{Margareta Ackerman \and David Loker}

\institute{Department of Computer Science, San Jose State University\\
San Jose, CA USA\\
\email{margareta.ackerman@sjsu.edu}\\
\and
Orbitwerks\\
San Jose, CA  USA\\
\email{dloker@gmail.com}}

\begin{document}
\maketitle

\begin{abstract}
This paper introduces ALYSIA: Automated LYrical SongwrIting Application.  ALYSIA is based on a machine learning model using Random Forests, and we discuss its success at pitch and rhythm prediction. Next, we show how ALYSIA was used to create original pop songs that were subsequently recorded and produced. Finally, we discuss our vision for the future of Automated Songwriting for both co-creative and autonomous systems. 
\end{abstract}

\section{Introduction}

Music production has traditionally relied on a wide range of expertise far exceeding the capabilities of most individuals. Recent advancements in music technology and the parallel development of electronic music has brought the joy of music-making to the masses. Among the countless amateurs who enjoy expressing their creativity through music-making, there has emerged a newly-leveled playing field for professional and semi-professional musicians who are able to independently produce music with no more than a personal computer. 

Technological advances have greatly simplified music production through systems such as Logic Pro\footnote{http://www.apple.com/logic-pro/}, Live\footnote{https://www.ableton.com/en/live/}, or even the freely available Garage Band\footnote{http://www.apple.com/mac/garageband/}. In addition to placing a wide range of instruments at our fingertips, without requiring the corresponding training, these systems provide a variety of mastering tools that enable the creation of high quality music with relative ease. Yet, despite this paradigm shift, the writing of songs - comprised of both melody and lyrics - remains an exclusive domain. A careful examination of electronic music reveals these limitations. The genre relies heavily on remixing existing music and much attention is paid to sound design, which does not require original composition or its juxtaposition with lyrics.

The creation of a new song continues to rely on diverse skills rarely possessed by one individual. The score alone calls for a combination of poetic ability to create the lyrics, musical composition for  the melody, and expertise in combining lyrics and melody into a coherent, meaningful and aesthetically pleasing whole. Once the score is complete, vocal skill and production expertise are needed to turn it into a finalized musical piece. While some may be satisfied with remixing existing music, the creation of truly original songs is often desired, in part due to restrictions associated with performing and remaking existing works. Can technology assist in the creation of original songs, allowing users to express their creativity through the skills they possess while letting technology handle the rest? 

We introduce ALYSIA, a songwriting system specifically designed to help both musicians and amateurs write  songs, with the goal of producing professional quality music. ALYSIA focuses on one of the most challenging aspects of the songwriting process, enabling the human creator to discover novel melodies to accompany lyrics. To the best of our knowledge, ALYSIA is the first songwriting system to utilize formal machine learning methodology. This lets us evaluate the success of our model based on its ability to correctly predict the pitch and rhythm of notes in the test set (which was not used to train the model).  On a corpus of pop music, the system achieves an impressive accuracy of 86.79\% on rhythms and 72.28\% on scale-degrees.

 In the generation phase, ALYSIA is given short lyrical phrases, for which it can then provide numerous melody options in the style of the given corpus. As it stands today, ALYSIA is a co-creative system that allows the user to retain a sense of creative control while eliminating one of the most essential skills traditionally required for songwriting: the ability to effectively explore the space of potential accompanying melodies. As a co-creative songwriting partner, ALYSIA assists human creators in the song-making progress by filling in gaps in the user's expertise. 
 

Helping people be creative, particularly in a manner that aids their social engagement both on and offline, can have substantial positive impact on their sense of well-being and happiness~\cite{gauntlett2013making}. Technological advancement has led to the radical expansion of those who are able to engage in music making, and made possible today's large online  communities for sharing musical works. Widening access to original songwriting stands not only to ease the path to professional musicianship, but also makes it possible for more people to enjoy the psychological benefits of engaging in this creative and socially rewarding activity.

Without any changes to the system, ALYSIA could be used to create complete melodies without human interference by simply selecting her first suggestion for each line of lyrics. However, a little human interaction can result in better music and gives the user creative freedom that would be lacking from even the best autonomous system. 









We begin with an examination of previous work and how it relates to our contributions, followed by a technical description of our system. We then move onto the application of ALYSIA to songwriting, starting with a detailed discussion of our proposed co-creative process. Finally, we showcase three musical pieces created with ALYSIA, and conclude with a discussion of our vision for the future of algorithmic songwriting. 




\section{Previous work}

Algorithmic composition has a rich history dating back to the 1950s. A wide range of techniques have been applied to music generation (with no lyrics), spanning from expert systems to neural networks.  See a survey by Fern{\'a}ndez and Vico~\cite{fernandez2013ai} for an excellent overview. 

Our focus here is on the creation of songs, which introduces a lyrical component to algorithmic composition. Algorithmic songwriting is relatively new, and so far the state of art addresses it using Markov chains. For example, M.U. Sucus-Apparatusf~\cite{toivanen2013automatical} by Toivanen et al, is a system that was used for the generation of Finnish art songs. Rhythm patterns are randomly chosen from among those typically found in art songs. Chord progressions are  subsequently generated by using second order Markov chains. Lastly, pitches are generated using a joint probability distribution based on chords and the previous note. M.U. Sucus-Apparatusf integrates the entire songwriting process, from lyric generation to melody and accompaniment. Yet, as with most previous systems, the major weakness is a lack of clear phrase structure \cite{toivanen2013automatical}. 


Another full-cycle automated songwriting system comes from the work of Scirea et al~\cite{scirea2015smug}. Their songwriting system SMUG was used to write songs from academic papers. The methodology relies on the evolution of Markov chains. It uses a corpus of mixed-genre popular songs, and integrates several rules, such as the high probability of rests at the ends of words. 

Monteith at al. \cite{monteith2012automatic} study the generation of melodic accompaniments. While utilizing a corpus, the system works by generating hundreds of corpus-driven random options guided by a few rules, and then chooses among them with an evaluation criteria that incorporates some musical knowledge. This process is applied for both the rhythm and melody generation. For example, for generating rhythms, the system produces 100 possible downbeat assignments. The text of each line is distributed across four measures, so four syllables are randomly selected to carry a downbeat. 

In other related works, Nichols~\cite{nichols2009lyric} studies a sub-component of the problem we address here, particularly lyric-based rhythm suggestion. He considers the set of all accompanying rhythms, and defines a fully rule-based function to evaluate them, via considerations such as rare word emphasis through strong beats. As future work, Nichols suggests solving the rhythm generation processes through machine learning techniques, which is one of our contributions here. Lastly,  in the work of Oliveira~\cite{oliveira2015tra}, the pipeline is inverted and lyrics are generated based on the rhythm of the provided melody. 

Our approach does not encode any musical rules and uses Random forests instead of Markov chains. The lack of directly encoded musical knowledge gives our system true genre independence. Musical genres are incredibly diverse, and rules that make sense for one genre can make it impossible to generate songs in another. For example, penalizing melodies with many repeating consecutive pitches is often a sensible choice, but it is unsuited to pop music, where such sequences are common. A  machine learning model that does not explicitly encode musical expertise not only applies to existing genres, but can also adapt to all future musical genres, which can differ from prior music in unpredictable ways.


Random forests offer some advantages over using Markov chains. One draw-back of Markov chains is that while they can model likelihoods of sequences, they do not encapsulate the ``why'' of one note being chosen over another after the same initial sequence. For example, if note A is the first note, why is the next note B most of the time? Perhaps it is only B most of the time if the key signature is C-major, and it is actually F\# most of the time under a different context. Such context can be discovered and used in systems like random forests given a suitable feature set. Additionally, we rely on machine learning methodology to formally evaluate our model, which has not been done in previous work on automated songwriting.  Another difference is that we rely on a co-creative process, which aids with our vision to create high quality music. Finally, ALYSIA is the first songwriting system whose songs were recorded and produced.

\section{The making of ALYSIA}

ALYSIA is an entirely data-driven system that is based upon a prediction model. We chose to construct two models. The first predicts note duration, which we refer to as the \emph{rhythm model}. The second predicts the scale degree of a note, with possible accidentals, which we refer to as the \emph{melody model}. Both models rely on a similar set of features, though depending on which model is run first, one model will have access to extra information. In our case, the melody model benefits from the predictions made by the rhythm model. Note that combing the models would increase the number of output classes, requiring a larger corpus. 


The most important component of our model is the set of features. The aim is to construct features that will allow the model to learn the mechanics and technicalities behind the structure of songs and their composition. 


Since we provide melodies for text provided by the user, our training set consists of vocal lines of music, and not arbitrary melodies. The corpus consists of Music-XML (MXL) files, with a single instrument corresponding to the vocal line and the accompanying lyrics. As such, each note has a corresponding syllable. 


The system consists of five distinct components.
\begin{enumerate}
\item The corpus
\item Feature extraction 
\item Model building and evaluation
\item User lyrics to features
\item Song generation
\end{enumerate}

We now discuss each of these in detail.

\subsection{Feature Extraction}

In order to build a model, we first extract a set of features from the Music-XML (MXL) files. 


For each note, we extract the following features:
\begin{itemize}
\item First Measure - A boolean variable indicating whether or not the current note belongs to the first measure of the piece
\item Key Signature - The key signature that applies to the current note
\item Time Signature - The time signature that applies to the current note
\item Offset - The number of beats since the start of the music
\item Offset within Measure - The number of beats since the start of the measure
\item Duration - The length of the note
\item Scale Degree - The scale degree of the note (1-7)
\item Accidental - The accidental of the note (flat, sharp, or none)
\item Beat Strength - The strength of beat as defined by music21. We include the continuous and categorical version (Beat Strength Factor) of this variable. 
\item Offbeat - A boolean variable specifying whether or not the note is offbeat
\item Syllable Type\textbf{*} - Classifies the current syllable in one of the following four categories: Single (the word consists of a single syllable), Begin (the current syllable is the first one in its word), Middle (the current syllable occurs in the middle of its word), End  (the current syllable is the last one in its word).
\item Syllable Number\textbf{*} - The syllable number within its word
\item Word Frequency\textbf{*} - The word frequency of the word which the current note/syllable is part of. This value is obtained through the dictionary frequency as calculated by Pythons NLTK (\href{http://www.nltk.org/}{http://www.nltk.org/}) library using the Brown corpus~\footnote{Brown corpus: \href{http://www.hit.uib.no/icame/brown/bcm.html}{http://www.hit.uib.no/icame/brown/bcm.html}}. 
\item Word Rarity\textbf{*} - A function of word frequency, as defined by \cite{nichols2009lyric}. $WordRarity = 2(1-\frac{\log_{10}(WordFrequency)}{7}).$
\item Word Vowel Strength\textbf{*} - The word vowel strength (primary, secondary, or none) as indicated by the  CMUDict v0.07.\footnote{CMUDict v0.07, Carnegie Mellon University.}
\item Number of Vowels in the Word\textbf{*} - The number of vowels found in the word corresponding to the current note/syllable. 
\item Scale Degree, Accidental, and Duration of previous 5 notes
\end{itemize}

The featured marked with an * are generated from lyrics found within the vocal line. These same features must be generated using the user's lyrics when ALYSIA is used in generation mode. The features marked with an * were used by \cite{nichols2009lyric} as part of a rule-based evaluation criteria for rhythm suggestion. 


\subsection{Model Building and Evaluation}

Using R, we train two models using random forests. Random forests were chosen for several reasons. They are well-suited for large numbers of categorical variables. Additionally, they allow non-linearity in combining features, without an explosion in the required size of the training set, in order to avoid overfitting.  Non-linearity makes random forests more powerful than linear models such as logistic regression. We randomly split the data using stratified sampling into a training set (75\% of the data) with the rest for testing.   

The accuracy of the rhythm model on test data was 86.79\%. Figure~\ref{fig:confRhythm} shows the confusion matrix. The row labels correspond to the duration found in the test sets, and column labels are the predicted labels. The number after the underscore corresponds to the number of dots in the durations. So, for example, a dotted sixteenth note is 16th\_1. 

As you can see in the confusion matrix, the model tends to perform better on rhythms which occur more frequently in the data set. A larger data set will enable us to test whether the trend will continue. 

\begin{figure*}
    \advance\leftskip-2cm
    \includegraphics[width = 16cm]{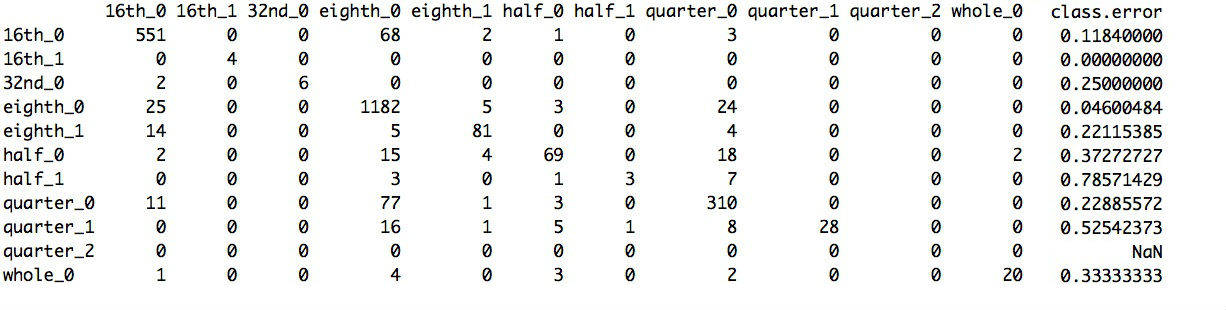}
    \caption{Confusion matrix for Rhythm Model. Cell ($i$,$j$) in the above table specifies how often a node with duration type $i$ was classified as having during $j$. The last column shows the error rate for each duration type (NaN indicating that no notes of this duration were found in the corpus).}
    \label{fig:confRhythm}
\end{figure*}


The accuracy of the melody model was 72.28\%. Figure~\ref{fig:confRhythm} shows the confusion matrix. The numbers correspond to the scale degrees, and the label after the underscore represents the accidentals. So, for example, 1\_none represents the first scale degree without accidentals. As with the rhythm model, we see that accuracy is higher when there are more examples. 

\begin{figure*}
    \advance\leftskip-1cm
    \includegraphics[angle=90, height = 22.5cm]{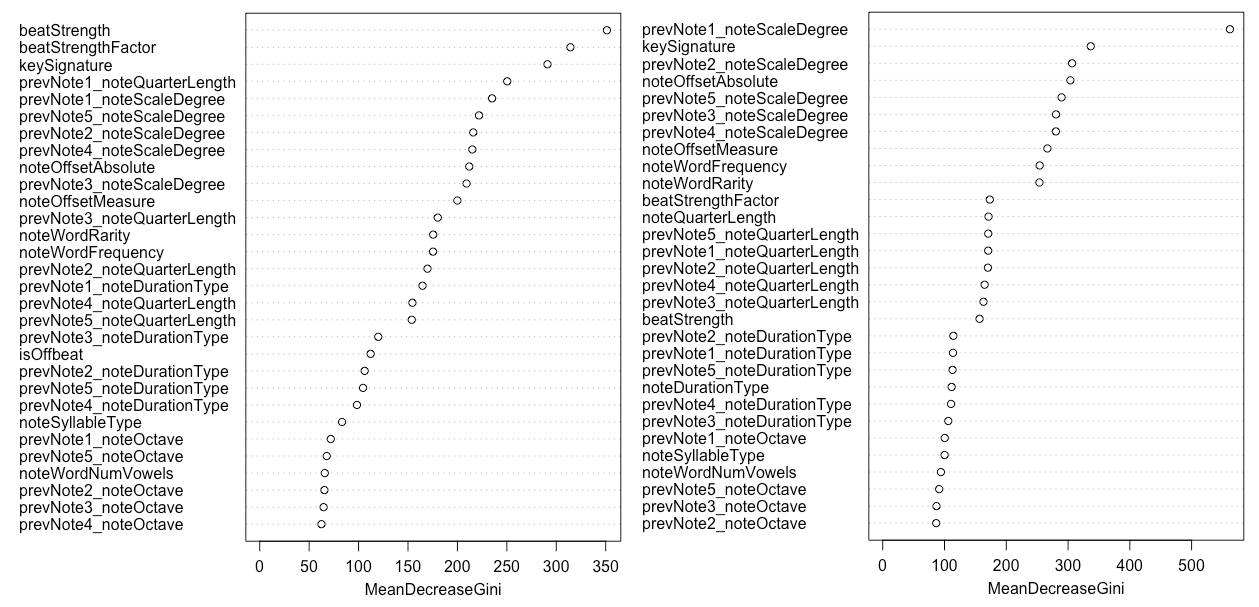}
    \caption{Importance plot for the rhythm model (left) and the melody model (right), representing the significance of each feature in the respective models.}
    \label{fig:modelDecAccuracy}
\end{figure*}

Lastly, our machine learning model allows us to evaluate the importance of each feature. Figure~\ref{fig:modelDecAccuracy} depicts the mean decrease Gini (which measures node impurity in tree-based classification) of each feature in the rhythm model (left) and the melody model (right). In both models, all 5 previous notes' scale degrees are some of the most important features. This suggests that including features for additional previous notes may be helpful. This also explains why Markov chains seem to do well for this task. However, we see that other features are also prominent. 

For example, the rhythm model is most heavily influenced by beat strength, revealing the potential of using a data-driven approach over Markov chains, which do not give access to such features. Finally, it is important to note that lyrical features such as word rarity and frequency also play an important role. 

\begin{figure*}
    \advance\leftskip-2cm
    \includegraphics[width = 16cm]{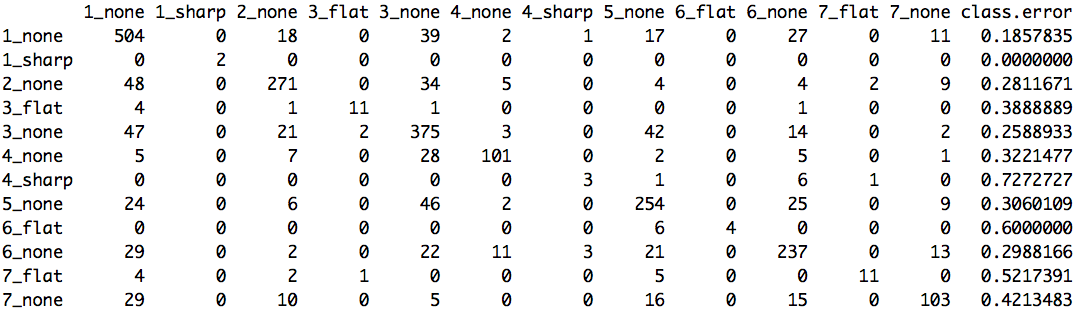}
    \caption{Confusion matrix for Melody Model. Cell ($i$,$j$) in the above table specifies how often scale degree $i$ was classified as scale degree $j$. The last column shows the error rate for each scale degree.}
    \label{fig:confRhythm}
\end{figure*}

\subsection{User Lyrics to Features} In order to generate music for a new set of lyrics using our models, we need to extract the same set of lyric features we were using for our models. This was done using a python script, which outputs a feature matrix for importing into R.

\subsection{Song Generation} Using R, we loop through the lyric feature set one sentence at a time. For each line, we read the feature set from the lyrics given by the user and generate the rhythm, followed by the melody. Note that we generate one note at a time for each syllable in the lyrical phrases. We also keep track of 5 previous notes, since these are very important features in the model. Finally, we generate a file detailing the generated melodies for each line. There may be many depending on how many melodies per line were requested.

\subsection{Corpus}
Our corpus consists of 12,358 observations on 59 features that we extract from MXL files of 24 pop songs. From each song, only the melody lines and lyrics were used. For model building and evaluation, the data is split using stratified sampling along the outcome variable, which is scale degree with accent for the melody model, and note duration for the rhythm model. Seventy-five percent of the data is used for training, while the remaining data is used for evaluation.

Please note that unlike MP3 files, MXL files containing lyrics are much more difficult to acquire, limiting the size of our data. Previous work on automated songwriting at times does not use any training (it is instead rule-based), or  trains separately on lyrics and melodies - which eliminates the possibility of learning how they interact. The size of the training set in previous work is often omitted\footnote{Due to the difficulties in attaining MXL files, we are currently creating a new larger corpus of songs, which will be analyzed in future work.}.

\subsection{Parameters} 
When used in generation mode, ALYSIA allows the user to tune the following parameters:

\subsubsection{Explore/Exploit} This parameter determines how heavily we lean on the model. Specifically, for each scale degree/note duration we generate, our model outputs a distribution over all possible outcomes. This parameter determines how many independent draws we make from this distribution. The final resulting note is the most common draw, with ties being broken by the original outcome distribution. That is, if scale degree 1 and 2 are tied after four draws, we take the scale degree that was originally more likely in the distribution output by the model for this data point. Given this definition, a higher explore/exploit parameter value means we exploit because we are almost always going to output the scale degree or duration that has the highest probability of occurring. This parameter allows us to favor the scale degrees and durations that are most likely, versus potentially taking a more varied approach and generating music that could be considered more experimental.

\subsubsection{Melody Count} We generate a set number of melodies per line, outputting them to a file for later translation into midi and MXL files.

\subsubsection{Rhythm Restriction} We allow the user to restrict the possible rhythmic outcome if they would like to omit certain note durations. For example, the user may want to disallow whole notes, or some faster notes such as a 32nd note.

\newpage
\section{The Co-Creative Process of Songwriting with ALYSIA}

In this section, we describe the co-creative process of songwriting with ALYSIA. This is a new approach to writing songs that requires minimal to no musical training. ALYSIA makes it easy to explore melodic lines, reducing songwriting to the ability to select melodies based on one's musical taste.

The user provides ALYSIA with the lyrics broken down into separate lines. Then, the system gives the specified number of melodies (notes/rhythm combinations) to which the given lyrics can be sang. The number of melodies per line is specified by the user. ALYSIA outputs the melodies in both MXL and MIDI format. The MIDI form is particularly helpful, since it allows the user to play the variations, select the desired one, and directly utilize the MIDI to produce the song. 

We now describe the workflow we used with ALYSIA. We typically ask for between 15 to 30 melodic variations per line of text. Among these, we select between three and ten melodic lines. It should be noted that nearly all of ALYSIA's suggestions are reasonable. For example, Figure~\ref{fig:variations} lists the first 12 melody options provided by ALYSIA for the phrase \emph{everywhere I look I find you looking back.} All of these melodies  work fairly well with the underlying text and could be extended into full songs. 

One may ask why we choose to look at 15 to 30 options if all are reasonable.  Having a variety of options can lead to better quality songs while also enabling artists to incorporate their own musical preferences, without possessing the composition and text/melody juxtaposition skills traditionally required to engage in this art form.

When making our selections, we look for melodies that are independently interesting, and have intriguing relationships with the underlying text. We search for lines that match the emotional meaning of the text (happy or sad), as well as interesting word emphasis. For example, if the word ``sunshine'' appears in the text, we may select a melody that rises with this word (see Figure~\ref{fig:Rainbows}). ALYSIA often suggests melodic variations for which we find interesting explanations. For example, in the original song \emph{Why Do I Still Miss You} (see Figure~\ref{fig:WhyDoIScore}), ALYSIA suggested a melody where the phrase ``went wrong'' is on the lower end of scale, giving these words an appropriately dark interpretation. 

The next step involves combining the melodic lines to form the complete song. This takes some trial and error, and several interesting variations can typically be attained. The most efficient approach we had found was to construct the song around a few melodic lines. Say, if there are particularly interesting melodies for lines 1 and 3, then lines 2 and 4 are chosen to fit with these. Having multiple options allows the artist to create several variations for verses and chorus repetitions, as often found in songs. 

\noindent%
\begin{minipage}{\linewidth}%
\makebox[\linewidth]{
    \includegraphics[width=11cm]{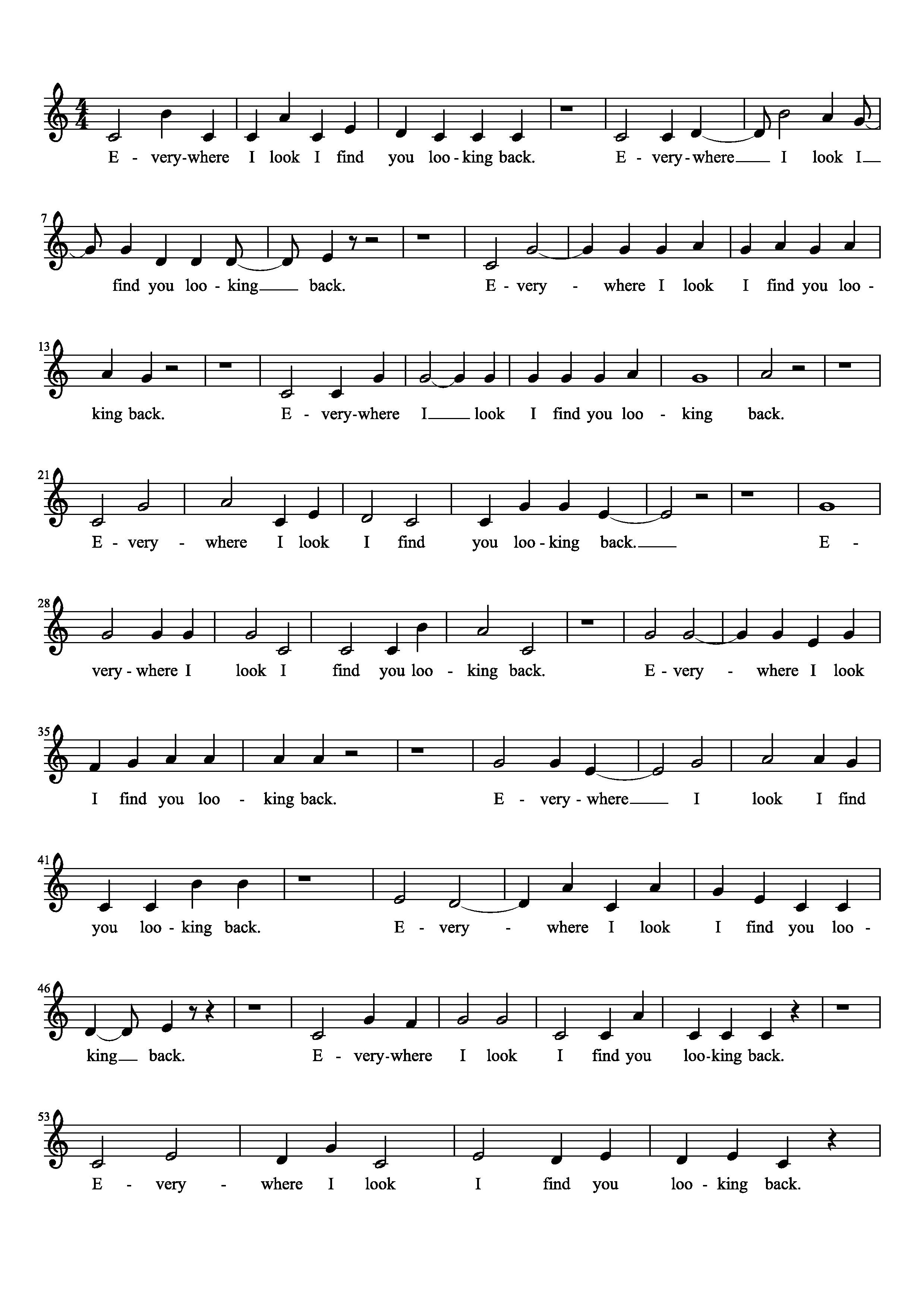}}
    \vspace{-15mm}
    \captionof{figure}{The first 12 melodies generated by ALYSIA to accompany the first line of the song \emph{Everywhere} (the full song appears in the following section). All melodies are sensible and could be used towards the creation of a complete melody line. Having many reasonable options allows the user to incorporate their musical taste into songwriting with ALYSIA.}
    \label{fig:variations}
\end{minipage}

\newpage
\section{Songs made with ALYSIA}

Three songs were created using ALYSIA. The first two rely on lyrics written by the authors, and the last borrows its lyrics from a 1917 Vaudeville song. The songs were consequently recorded and produced. All recordings are posted on \href{http://bit.ly/2eQHado}{http://bit.ly/2eQHado}.

The primary technical difference between these songs was the explore/exploit parameter. \emph{Why Do I Still Miss You} was made using an explore vs exploit parameter that leads to more exploration, while \emph{Everywhere} and \emph{I'm Always Chasing Rainbows} were made with a higher exploit setting.


\subsection{\emph{Why Do I Still Miss You}}

\emph{Why Do I Still Miss You}, the first song ever written with ALYSIA, was made with the goal of testing the viability of using computer-assisted composition for production-quality songs. The creation of the song started with the writing of original lyrics, written by the authors.  The goal was to come up with lyrics that stylistically match the style of the corpus, that is, lyrics in today's pop genre (See Figure~\ref{fig:WhyDoIScore}). 







Next, each lyrical line was given to ALYSIA, for which she produced between fifteen and thirty variations. In most cases, we were satisfied with one of the options ALYSIA proposed within the first fifteen tries.  We then combined the melodic lines. See Figure~\ref{fig:WhyDoIScore} for the resulting score. Subsequently, the song was recorded and produced. One of the authors, who is a professionally trained singer, recorded and produced the song. A recording of this song can be heard at \href{http://bit.ly/2eQHado}{http://bit.ly/2eQHado}. 

\vspace{4mm}
\noindent%
\begin{minipage}{\linewidth}%
\makebox[\linewidth]{
    \includegraphics[width=10cm]{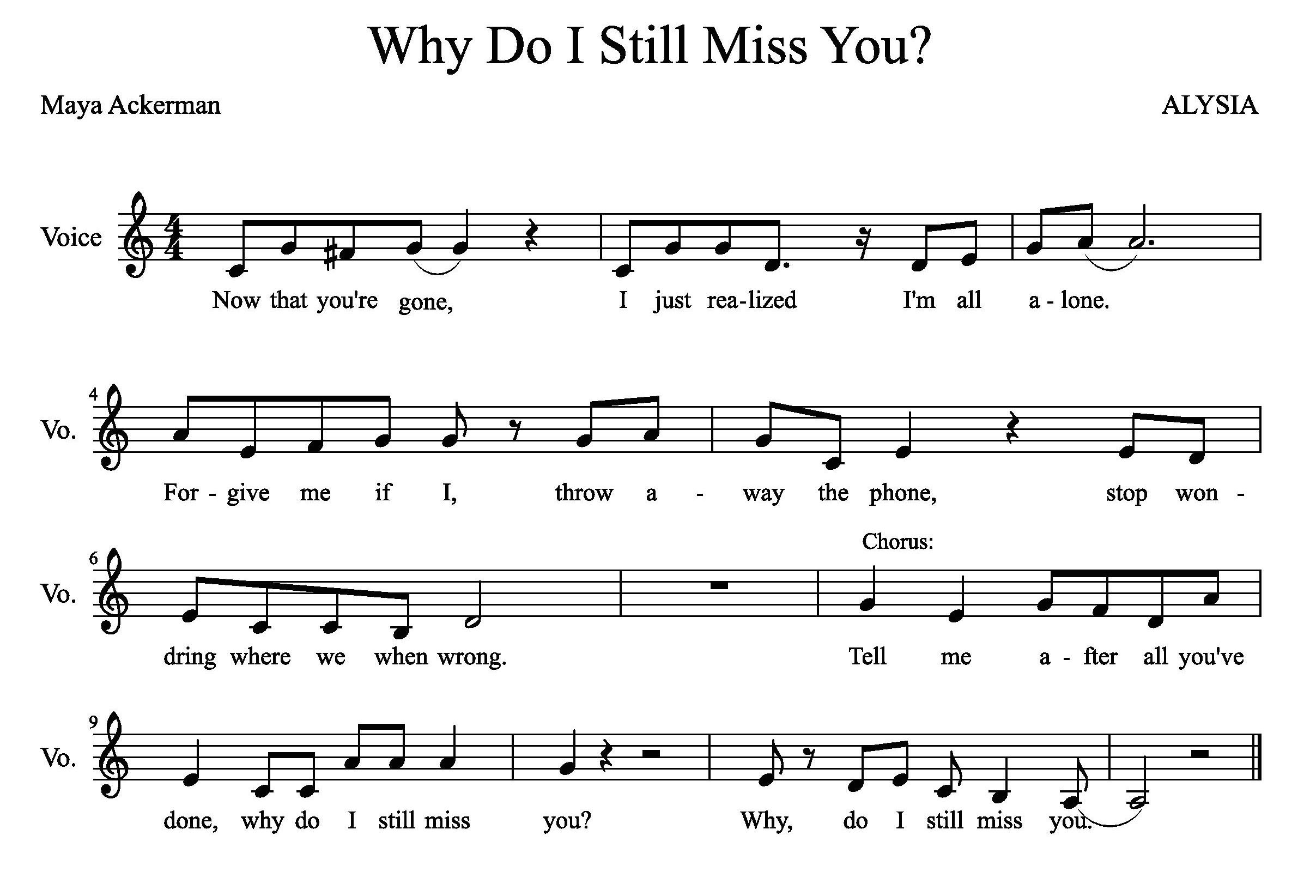}}
    \vspace{-3mm}
    \captionof{figure}{An original song co-written with ALYSIA.}
    \label{fig:WhyDoIScore}
\end{minipage}
\vspace{2mm}



This experiment, intentionally spanning the entire songwriting process from the lyrics to the final production, demonstrates the viability of our approach for making production-worthy works. It also shows that ALYSIA makes songwriting accessible to those with no songwriting ability, as the song was written by one of the authors, who does not possess this skill unassisted by our computer system. Informal assessment by the authors and many others consistently point to the surprising quality of this song, gauging it as having quality on par with that of fully human-made songs. Future user studies will help to formally assess this claim. 

For this song, we used an explore/exploit parameter of 1, meaning that the distribution of notes output by the model was sampled only once. With regard to improving the application of our system, the making of \emph{Why Do I Still Miss You} suggested that we should increase the exploit parameter. This was particularly important for the rhythm suggestion, which we found to be too varied. As such, \emph{Why Do I Still Miss You} fully utilized the melody model, while making some manual adjustment to the rhythm. We stress that even with this sub-optimal setting for the exploit-explore setting, ALYSIA made songwriting accessible to someone who previously could not engage in this artform.  

The two subsequent songs discussed below exploit our model more heavily, using sampling four times for the rhythm model and two times for the melody model, selecting the majority, which corresponds to an explore/exploit parameter of 4 and 2, respectively. Heavier reliance on the model allowed us to use the rhythm suggestions without alterations and also raised the proportion of high quality scale degree sequences.

\subsection{\emph{Everywhere}}\label{sec:Everywhere}

\emph{Everywhere} is the first original song written with ALYSIA using a high setting of the exploit parameter. This made it even easier to discover new melodies. In fact, so many of ALYSIA's melodies were interesting that combining melodies became more time consuming (of course, one could simply consider a smaller set of melodies per line to overcome this challenge). Nevertheless, the writing of \emph{Everywhere} from start to finish took only about two hours. Figure~\ref{fig:everywhere} shows the score co-created with ALYSIA. A recording appears on \href{http://bit.ly/2eQHado}{http://bit.ly/2eQHado}






 \subsection{\emph{I'm Always Chasing Rainbows}}


The final song made with ALYSIA so far relies on lyrics from the song \emph{I'm Always Chasing Rainbows}. The original music is credited to Harry Carroll, although the melody was in fact adapted from Chopin's Fantaisie-Impromptu. Lyrics were written by  Joseph McCarthy. The song's appearance in Broadway musicals and motion pictures contributed to its popularity. As it was published in 1917, \emph{I'm Always Chasing Rainbows} is in public domain. 

Due to its Vaudeville's style and classical roots, we thought it would be interesting to recreate \emph{I'm Always Chasing Rainbows} as a pop song. ALYSIA's version is given in Figure~\ref{fig:Rainbows}. A production of ALYSIA's pop version of this Vaudeville song has been posted on \href{http://bit.ly/2eQHado}{http://bit.ly/2eQHado}, where you can also find the original score. 

\noindent%
\begin{minipage}{\linewidth}%
\makebox[\linewidth]{
    \includegraphics[width=9cm]{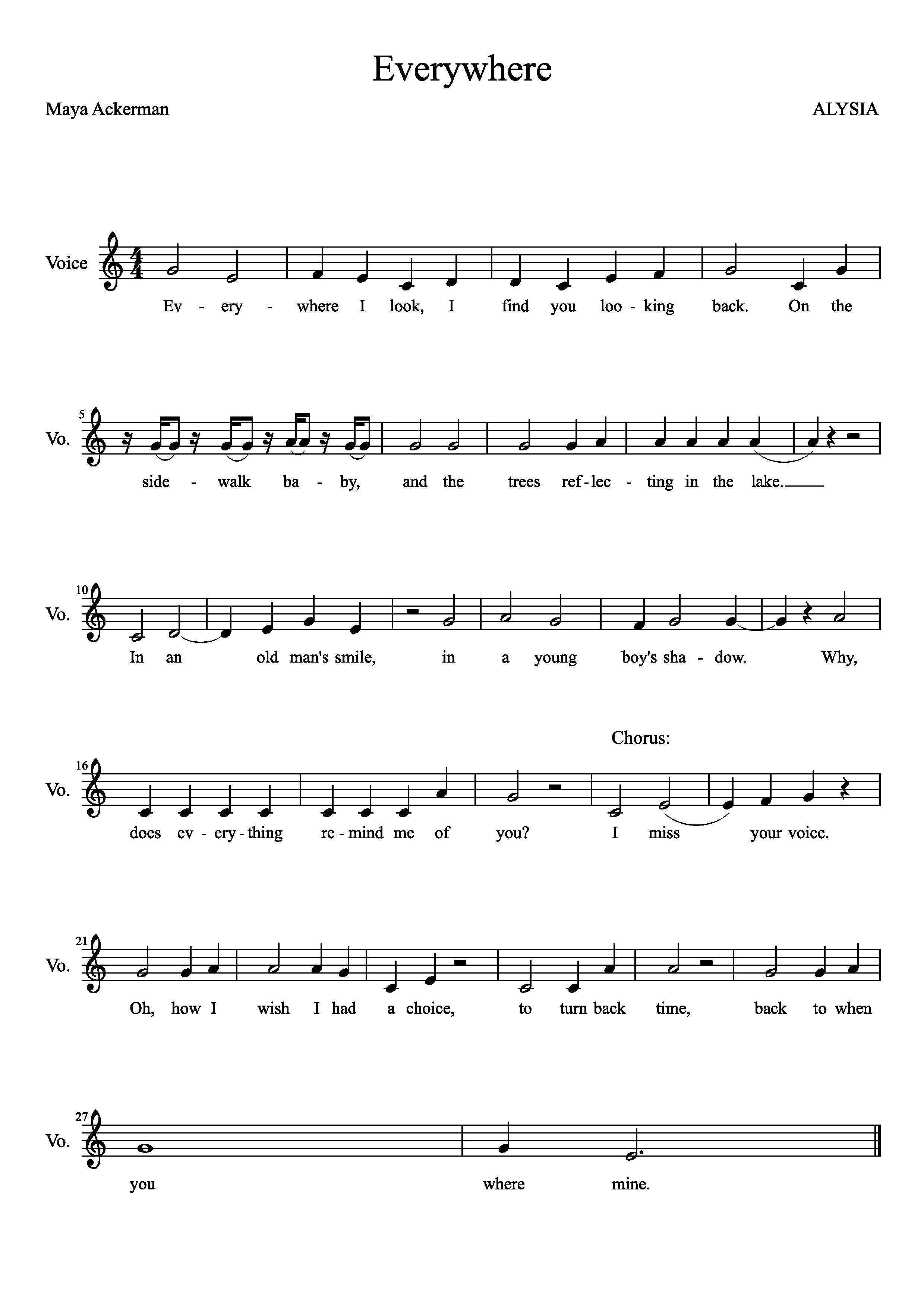}}
    \vspace{-15mm}
    \captionof{figure}{An original song co-written with ALYSIA.}
    \label{fig:everywhere}
\end{minipage}

\section{Discussion: Co-Creative and Autonomous Songwriting}


Songwriting is the art of combining melodies and lyrics; It is not enough to have beautiful melodies and poetic lyrics, the music and words must fit together into a coherent whole. This makes Algorithmic Songwriting a distinct sub-field of Algorithmic Composition. ALYSIA is the first machine-learning system that learns the relationship between melodies and lyrics, and uses the resulting model to create new songs in the style of the corpus. 

The unique challenges of songwriting were observed during the creation of the first musical to rely on computational creativity systems, Beyond the Fence, which played in London in the early months of 2016~\cite{jordanous2016has}. During the panel on  Beyond the Fence held at the Seventh International Conference of Computational Creativity, it was noted that the juxtaposition of music and text posed a substantial challenge in the making of this musical. This further illustrates the need for systems that focus on uncovering the complex relationships between music and lyrics.

\vspace{-1mm}
\noindent%
\begin{minipage}{\linewidth}%
\makebox[\linewidth]{
    \includegraphics[width=9cm]{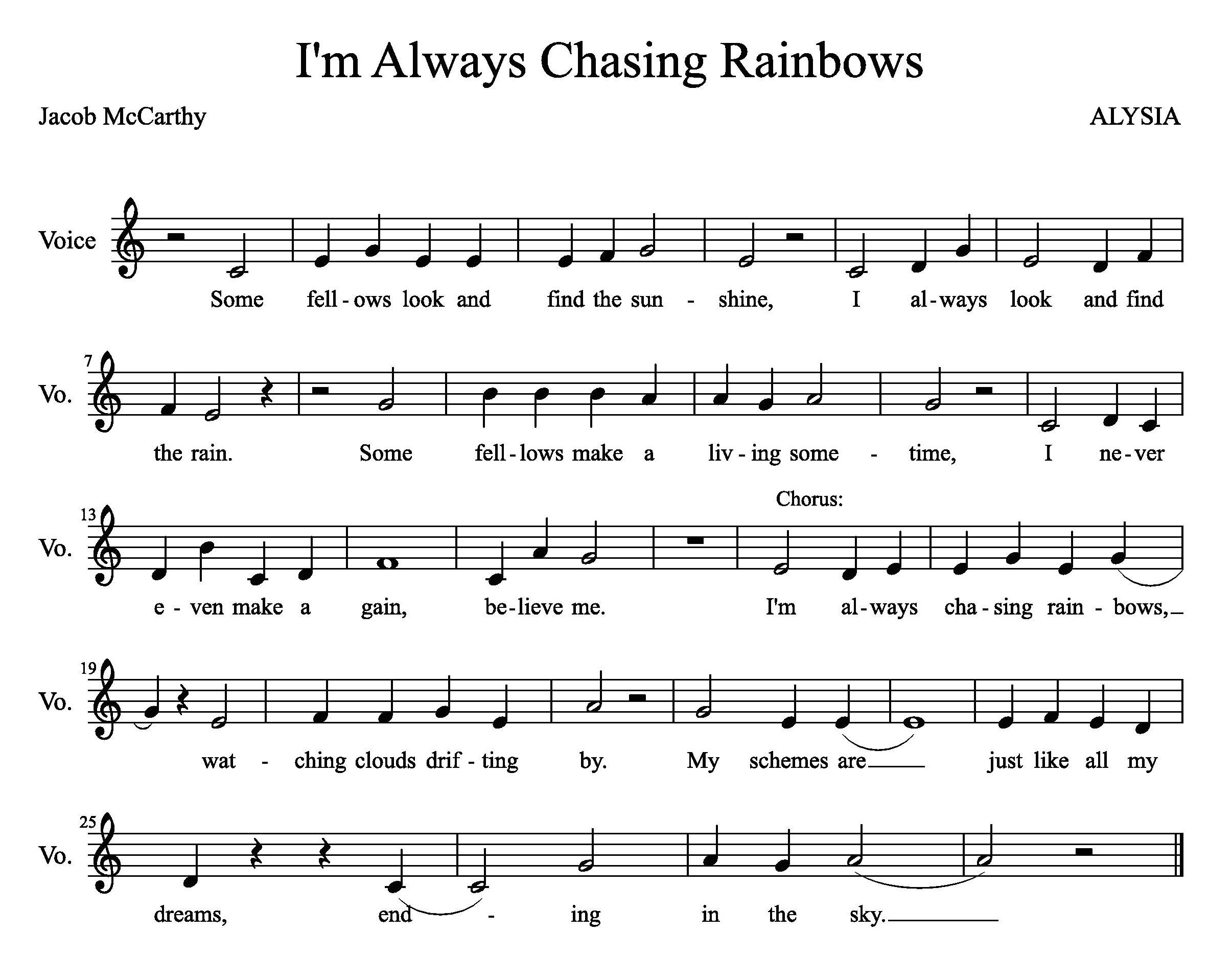}}
    \captionof{figure}{Melody and rhythm by ALYSIA, lyrics by Joseph McCarthy from the popular Vaudeville song \emph{I'm Always Chasing Rainbows}.}
\label{fig:Rainbows}
\end{minipage}
\vspace{2mm}
   

Algorithmic songwriting offers intriguing challenges as both an autonomous and a co-creative system. An autonomous songwriting system producing works on par with those of expert human songwriters would mark a significant achievement. Yet, we can go beyond the score. What if, in addition to writing the song, an automated system could also perform and record its own compositions? A truly independent system would not only create a score, but incorporate the full spectrum of expertise required for the creation of a complete song, including the vocal performance, expressive rendition, and automated music production.


As we aspire to create autonomous songwriters, artists and hobbyists alike are thirsty for our help. 
Even if we had access to a fully autonomous songwriter, it would not replace the need for a corresponding co-creative system. Whereas an autonomous songwriter could be used when complete works are desired, a co-creative variation would satisfy the human need for music making - much like the difference between the joys of eating and cooking. A co-creative algorithmic songwriter would expand our creative repertoire, making songwriting accessible to those who cannot otherwise enjoy this art-form, and be of great use to semi-professional and amateur musicians (particularly singers) who do have the luxury of hiring a songwriter. 

 In the development of a co-creative songwriting system, it is desired that the users retain creative control, allowing them to claim ownership of the resulting works, or at least experience the process as an expression of creativity. The goal is to relieve the burden of having to master all of the diverse skills needed for the creation of a song, giving users the freedom to focus on aspects of the creative process in which they either specialize or find most enjoyable. 
 


From a technical viewpoint, the goals of autonomous and co-creative algorithmic songwriters converge - the primary difference being that a co-creative system allows for more human interaction. ALYSIA is a long-term project in which we will simultaneously explore both of these paths as the system's technical foundation continues to expands.



Short-term goals for ALYSIA include the automation of chord progressions that would underlie the melody. A couple of user studies are being planned where people at different levels of musical expertise will evaluate the songs made with ALYSIA. To further our machine learning methodology, we are putting together a corpora of songs in Music XML format. For several different genres, a large corpus will be used to train a model using a neural network, which requires large data sets to avoid over-fitting. This will also let us explore differences in melodies resulting from different machine learning models.  We are also experimenting with our system's potential for songwriting in different genres and languages, ranging from English pop songs discussed here to the creation of new arias in the style of classical Italian composer Giacomo Puccini. 

\vspace{-2mm}
\bibliographystyle{plain}
\bibliography{references}

\begin{thebibliography}{1}

\bibitem{fernandez2013ai}
Jose~D Fern{\'a}ndez and Francisco Vico.
\newblock Ai methods in algorithmic composition: A comprehensive survey.
\newblock {\em Journal of Artificial Intelligence Research}, pages 513--582,
  2013.

\bibitem{gauntlett2013making}
David Gauntlett.
\newblock {\em Making is connecting}.
\newblock John Wiley \& Sons, 2013.

\bibitem{jordanous2016has}
Anna Jordanous.
\newblock Has computational creativity successfully made it ‘beyond the
  fence’in musical theatre?
\newblock In {\em Proceedings of the 7th International Conference on
  Computational Creativity}, 2016.

\bibitem{monteith2012automatic}
Kristine Monteith, Tony Martinez, and Dan Ventura.
\newblock Automatic generation of melodic accompaniments for lyrics.
\newblock In {\em Proceedings of the International Conference on Computational
  Creativity}, pages 87--94, 2012.

\bibitem{nichols2009lyric}
Eric Nichols.
\newblock {\em Lyric-based rhythm suggestion}.
\newblock Ann Arbor, MI: Michigan Publishing, University of Michigan Library,
  2009.

\bibitem{oliveira2015tra}
Hugo~Gon{\c{c}}alo Oliveira.
\newblock Tra-la-lyrics 2.0: Automatic generation of song lyrics on a semantic
  domain.
\newblock {\em Journal of Artificial General Intelligence}, 6(1):87--110, 2015.

\bibitem{scirea2015smug}
Marco Scirea, Gabriella~AB Barros, Noor Shaker, and Julian Togelius.
\newblock Smug: Scientific music generator.
\newblock In {\em Proceedings of the Sixth International Conference on
  Computational Creativity June}, page 204, 2015.

\bibitem{toivanen2013automatical}
Jukka~M Toivanen, Hannu Toivonen, and Alessandro Valitutti.
\newblock Automatical composition of lyrical songs.
\newblock In {\em The Fourth International Conference on Computational
  Creativity}, 2013.

\end{thebibliography}

\end{document}